\documentclass[draftclsnofoot,onecolumn,11pt]{IEEEtran}

\usepackage{amsmath,amssymb,enumerate,subfigure,multirow,hhline,color}
\usepackage{xcolor}
\usepackage{hyperref}
\usepackage{graphicx}
\usepackage{graphics}
\usepackage[sort]{cite}
\usepackage{amsthm}
\usepackage{algorithm}
\usepackage{algpseudocode}
\usepackage{enumitem}

\usepackage{epsfig,psfrag}
\usepackage{tabularx}
\usepackage{tabulary}

\usepackage[sort]{cite}

\usepackage{hyperref}
\hypersetup{breaklinks=true,colorlinks=true,linkcolor=blue,citecolor=blue}

\usepackage[noabbrev,capitalise,nameinlink]{cleveref}

\DeclareMathOperator{\argmax}{arg\,max}

\newtheorem{example}{Example}

\allowdisplaybreaks
\usepackage{authblk}

\begin{document}

\title{Optimization for Reinforcement Learning: \\ From Single Agent to Cooperative Agents}

\author[1]{Donghwan Lee}
\author[1]{Niao He}
\author[2]{Parameswaran Kamalaruban}
\author[2]{Volkan Cevher}
\affil[1]{University of Illinois at Urbana-Champaign (UIUC)}
\affil[2]{\'Ecole Polytechnique F\'ed\'erale de Lausanne (EPFL)}

\maketitle
\begin{abstract}
This article reviews recent advances in multi-agent reinforcement learning algorithms for large-scale control systems and communication networks, which learn to communicate and cooperate. We provide an overview of this emerging field, with an emphasis on the decentralized setting under different coordination protocols. We highlight the evolution of reinforcement learning algorithms from single-agent to multi-agent systems, from a distributed optimization perspective, and conclude with future directions and challenges, in the hope to catalyze the growing synergy among distributed optimization, signal processing, and reinforcement learning communities.
\end{abstract}

\section{Introduction}
Fueled with recent advances in deep neural networks, reinforcement learning (RL) has been in the limelight for many recent breakthroughs in artificial intelligence, including defeating humans in games (e.g., chess, Go, StarCraft), self-driving cars, smart home automation, service robots, among many others. Despite these remarkable achievements, many basic tasks can still elude a single RL agent. 
Examples abound from multi-player games, multi-robots, cellular antenna tilt control, traffic control systems, smart power grids to network management.


Often, cooperation among multiple RL agents is much more critical: multiple agents must collaborate to complete a common goal, expedite learning, protect privacy, offer resiliency against failures and adversarial attacks, and overcome the physical limitations of a single RL agent behaving alone. These tasks are studied under the umbrella of \emph{cooperative multi-agent RL (MARL)}, where agents seek to learn optimal policies 
to maximize a shared team reward, while interacting with an unknown stochastic environment and with each other. Cooperative MARL is far more challenging than the single-agent case due to: i) the exponentially growing search space, ii) the non-stationary and unpredictable environment caused by the agents' concurrent yet heterogeneous behaviors, and iii) the lack of \textit{central} coordinators in many applications. These difficulties can be alleviated by appropriate coordination among agents. 

{ The cooperative MARL can be further categorized into subclasses depending on the information structure and types of coordination, such as how much information (e.g., state, action, reward, etc.) is available for each agent, what kinds of information can be shared among the agents, and what kinds of protocols (e.g., communication networks, etc.) are used for coordination. When only local partial state observation is available for each agent, the corresponding multi-agent systems are often described through decentralized partially observable Markov decision processes (MDP), or DEC-POMDP for short, for which the decision problem is known to be extremely challenging. In fact, even the planning problem of DEC-POMDPs (with known models) is known to be NEXT-complete~\cite{bernstein2002complexity}. Despite some recent empirical successes~\cite{leibo2017multi,foerster2017stabilising,foerster2016learning}, finding an exact solution of Dec-POMDPs using RLs with theoretical guarantees remains an open question.

When full state information is available for each agent, we call agents \emph{joint action learners} (JALs) if they also know the joint actions of other agents, and \emph{independent learners} (ILs) if agents only know their own actions. Learning tasks for ILs are still very challenging, since each agent sees other agents as parts of the environment, so without observing the internal states, including other agents’ actions, the problem essentially becomes non-Markovian~\cite{laurent2011world} and a partially observable MDP (POMDP). It turns out that optimal policy can be found under restricted assumptions such as deterministic MDP~\cite{matignon2012independent}, and for general stochastic MDPs, several attempts have demonstrated empirical successes~\cite{tampuu2017multiagent,lauer2000algorithm,lauer2004reinforcement}. For a more comprehensive survey on independent MARLs, the reader is referred to the survey~\cite{matignon2012independent}.

The form of rewards, either centralized or decentralized, also makes a huge difference in multi-agent systems. If every agent receives a common reward, the situation becomes relatively easy to deal with. For instance, JALs can perfectly learn exact optimal policies of the underlying decision problem even without coordination among agents~\cite{Claus1998thedynamics}. The more interesting and practical scenario is when rewards are decentralized, i.e., each agent receives its own local reward while the global reward to be maximized is the sum of local rewards. This decentralization is especially important when taking into account the privacy and resiliency of the system. 

Clearly, learning without coordination among agents is impossible under decentralized rewards. This article focuses on this important subclass of cooperative MARL with decentralized rewards, assuming the full state and action information is available to each agent. In particular, we consider decentralized coordination through network communications characterized by graphs, where each node in the graph represents each agent and edges connecting nodes represent communication between them.

Distributed optimization rises to the challenge by achieving global consensus on the optimal policy
through only local computation and communication with neighboring agents. Recently, several important advances have been made in this direction such as the distributed TD-learning~\cite{doan2019convergence}, distributed Q-learning~\cite{kar2013cal}, distributed actor-critic algorithm~\cite{zhang2018fully}, and other important results~\cite{zhang2018finite,qu2019value,wai2018multi,lee2018stochastic}. These works largely benefit from the synergistic connection between RLs and the core idea of averaging consensus-based distributed optimization~\cite{nedich2015convergence}, which leverages averaging consensus protocols for information propagation over networks and rich theory established in this field during the last decade.

In this survey, we provide an overview of this emerging field with an emphasis on optimization within the decentralized setting (decentralized rewards and decentralized communication protocols). For this purpose, we highlight the evolution of RL algorithms from single-agent to multi-agent systems, from a distributed optimization perspective, in the hope to catalyze the growing synergy among distributed optimization, signal processing, and RL communities. 

In the sequel, we first revisit the basics of single-agent RL in Section~II and extend to multi-agent RL in Section~III. In Section~IV, we provide preliminaries of distributed optimization as well as consensus algorithms. In Section~V, we discuss several important consensus-based MARL algorithms with decentralized network communication protocols. Finally, in Section~VI, we conclude with future directions and open issues. Note that our review is not exhaustive given the magazine limits; we suggest the interested reader to further read~\cite{bu2008comprehensive,matignon2012independent,nguyen2018deep}.
}


\section{Single-agent RL basics}
{ To understand MARL, it is imperative that we briefly review the basics of single-agent RL setting, where only a single agent interacts with an unknown stochastic environment. Such environments are classically represented by a Markov decision process: ${\cal M}:= ({\cal S},{\cal A},P,r,\gamma)$, where the state-space ${\cal
S}:=\{ 1,2,\ldots,|{\cal S}|\}$ and action-space ${\cal A}:= \{1,2,\ldots,|{\cal A}|\}$, upon selecting an action $a \in {\cal A}$ with the current state $s \in {\cal S}$, the state transits to $s'\in {\cal S}$ according to the state transition probability $P(s'|s,a)$, and the transition incurs a random reward $r(s,a)$. For simplicity, we consider the infinite-horizon (discounted) Markov decision problem (MDP), where the agent sequentially takes actions to maximize cumulative discounted rewards. The goal is to find a deterministic optimal policy, $\pi^*:{\cal S}\to {\cal A} $, such that 
\begin{align}
\pi^*:= \argmax_{\pi\in \Theta} {\mathbb E}\left[\sum_{k=0}^\infty {\gamma^k r(s_k,\pi(s_k))}\right],\label{eq:rl-single-objective}
\end{align}
where $\gamma \in [0,1)$ is the discount factor, $\Theta$ is the set of all admissible deterministic policies, and $(s_0,a_0,s_1,a_1,\ldots)$ is a state-action trajectory generated by the Markov chain under policy $\pi$. Solving MDPs involves two key concepts associated with the expected return:
\begin{enumerate}
\item $V^{\pi}(s):={\mathbb E}\left[\sum_{k=0}^\infty {\gamma^k r(s_k,\pi(s_k))}|s_0=s\right]$ is called the (state) value function for a given policy $\pi$, which encodes the expected cumulative reward when starting in the state $s$, and then, following the policy $\pi$ thereafter.
\item $Q^{\pi}(s,a):= {\mathbb E} \left[\sum_{k=0}^\infty {\gamma^k r(s_k,\pi(s_k))}|s_0=s,a_0=a\right]$ is called the state-action value function or Q-function for a given policy $\pi$, which measures the expected cumulative reward when starting from state $s$, taking the action $a$, and then, following the policy $\pi$. 
\end{enumerate}

Their optima over all possible policies are defined by $V^*(s):=\max_{\pi:{\cal S} \to {\cal A}} V^{\pi}(s) = \max_{a} Q^*(s,a)$ and $Q^*(s,a):=\max_{\pi:{\cal S} \to {\cal A}} Q^{\pi}(s,a)$, respectively. Given the optimal value functions $Q^*$ or $V^*$, the optimal policy $\pi^*$ can be obtained by picking an action $a$ that is greedy with respect to $V^*$ or $Q^*$, i.e., $\pi^*(s) = \argmax_{a} {\mathbb E}_{s' \sim P(\cdot|s,a)} [r(s,a) + \gamma V^* (s')]$ or $\pi^*(s) = \argmax_{a} Q^* (s,a)$, respectively. When the MDP instance, ${\cal M}$, is known, then it can be solved efficiently via dynamic programming (DP) algorithms. Based on the Markov property, the value function $V^\pi$ for a given policy $\pi$, satisfies the Bellman equation: $V^\pi(s)= {\mathbb E}_{s'\sim P(\cdot|s,\pi(s))} \left[r(s,\pi(s))+\gamma V^\pi (s')\right]$. The similar property holds for $Q^\pi$ as well. Moreover, the optimal Q-function $Q^*$, satisfies the \emph{Bellman optimality equation},  $Q^*(s,a)= {\mathbb E}_{s'\sim P(\cdot|s,a)} \left[r(s,a)+\max_{a'}\gamma Q^* (s',a')\right]$. Various DP algorithms, such as the policy and value iterations, are obtained by turning the Bellman equations into update rules.
}

\subsection{Classical RL Algorithms}
Many classical RL algorithms can be viewed as stochastic variants of DPs. This insight will be key for scaling MARL in the sequel. The temporal-difference (TD) learning is a fundamental RL algorithm to estimate the value function of a given policy $\pi$ (called as policy evaluation method): 
\begin{align}
V_{k+1}(s_k) = V_k(s_k)+ \alpha_k (r(s_k,\pi(s_k)) + \gamma V_k(s_{k+1}) - V_k(s_k)), \label{eq:TD}
\end{align}
where $s_k \sim d^\pi$, $s_{k+1} \sim P(\cdot|s_k,\pi(s_k))$, $d^\pi$ denotes the stationary state distribution under policy $\pi$, and $\alpha_k$ is the learning rate (or step-size). 
For any fixed policy $\pi$, TD update converges to $V^\pi$ almost surely (i.e., with probability $1$) if the step-size satisfies the so-called {\em Robbins-Monro rule}, $\sum_{k=0}^\infty
{\alpha_k}=\infty$, $\sum_{k=0}^\infty{\alpha_k^2}<\infty$~\cite{sutton1998reinforcement}. Although theoretically sound, the naive TD learning is only applicable to small-scale problems as it needs to store and enumerate values of all states. However, most practical problems we face in the real-world have large state-space. In such cases, enumerating all values in a table is numerically inefficient or even intractable.

Using function approximations resolves this problem by encoding the value function with a parameterized function class, $V(\cdot) \cong V(\cdot;\theta)$. The simplest example is the linear function approximation, $V(\cdot;\theta) = \Phi \theta$, where $\Phi = [\phi(1) ; \cdots ; \phi(|{\cal S}|)]^\top \in {\mathbb R}^{|{\cal S}| \times n}$ is a feature matrix, and $\phi:{\cal S}\to {\mathbb R}$ is a pre-selected feature mapping. TD learning update with linear function approximation is written as follows
\begin{align}
\theta_{k+1} = \theta_k + \alpha_k (r(s_k,\pi(s_k))+\gamma\phi(s_{k+1})^T \theta_k -\phi(s_k)^T \theta_k)\phi(s_k).\label{eq:TD_LFA}
\end{align}
The above update is known to converge to $\theta^*$ almost surely~\cite{tsitsiklis1997analysis}, where $\theta^*$ is the solution to the {\em projected Bellman equation}, provided that the Markov chain with transition matrix $P^{\pi}$ (state transition probability matrix under policy $\pi$) is ergodic and the step-size satisfies the Robbins-Monro rule. Finite sample analysis of the TD learning algorithm is only recently established in~\cite{bhandari2018finite,dalal2018finite,srikant2019}. Besides the standard TD, there also exits a wide spectrum of TD variants in the literature~\cite{sutton2009fast,sutton2009convergent,dai2017learning,lee2019target}. Note that when a nonlinear function approximator, such as neural networks, is used, these algorithms are not guaranteed to converge.

The policy optimization methods aim to find the optimal policy $\pi^*$ and broadly fall under two camps, with one focusing on value-based updates, and the other focusing on direct policy-based updates. There is also a class of algorithms that belong to both camps, called actor-critic algorithms. Q-learning is one of the most representative valued-based algorithms, which obeys the update rule
\begin{align}
Q_{k+1}(s_k,a_k)= Q_k(s_k,a_k) + \alpha_k (r(s_k,a_k) + \gamma \max_{a\in {\cal A}} Q_k(s_{k+1},a) - Q_k(s_k,a_k)), \label{eq:Q-learning}
\end{align}
where $s_k \sim d^\pi$, $s_{k+1} \sim P(\cdot|s_k,\pi^b(s_k))$, and $\pi^b$ is called the behavior policy, which refers to the policy used to collect observations for learning. The algorithm converges to $Q^*$ almost surely~\cite{bertsekas1996neuro} provided that the step-size satisfies the Robbins-Monro rule, and every state is visited infinitely often. { 
Unlike value-based methods, direct policy search methods optimize a parameterized policy $\pi_{\theta}$ from trajectories of the state, action, reward, $(s,a,r)$ without any value function evaluation steps, using the following (stochastic) gradient steps:
\begin{align}
\theta_{k+1}= \theta_k+\alpha_k \hat \nabla_\theta J(\theta_k), \text{ where } J(\theta):= {\mathbb E} \left[\sum_{k=0}^\infty {\gamma^k r_{\pi_\theta}(s_k)}\right], \label{eq:policy-search}
\end{align}
where $\hat \nabla_\theta J(\theta_k)$ is a stochastic estimate of the gradient evaluated at $\theta_k$. The gradient of the value function has the simple analytical form
$
\nabla J(\theta) = \mathbb{E}_{s\sim d_{\pi_\theta},a\sim\pi_\theta}[ \nabla \log\pi_\theta(a|s)Q^{\pi_\theta}(s,a)],
$
which, however, needs an estimate of the Q-function, $Q^{\pi_\theta}(s,a)$. The simple policy gradient method replaces $Q^{\pi_\theta}(s,a)$ with a Monte Carlo estimate, which is called REINFORCE~\cite{williams1992simple}. However,  the \textit{high variance} of the stochastic gradient estimates due to the Monte Carlo procedure often leads to slow and sometimes unstable convergence. The actor-critic methods combine the advantages of value-based and direct policy search methods~\cite{konda2003onactor} to reduce the variance. These algorithms parameterize both the policy and the value functions, and simultaneously update both in training
\begin{align*}
&{\rm Critic\,\, update}:w_{k+1}=w_k+\alpha_k (r(s_k,a_k)+\gamma Q(s_{k+1},a_{k+1};w_k)-Q(s_k,a_k;w_k))\nabla_w Q(s_k,a_k;w_k)\\
&{\rm Actor\,\,update}:\theta_{k+1}=\theta_k +\beta_k Q(s_k,a_k;w_k)\nabla_\theta\log\pi(a_k|s_k;\theta_k),
\end{align*}
where $w_k$ and $\theta_k$ are parameters of the value and policy, respectively.  They often exhibit better empirical performance than value-based or direct policy-based methods alone. Nonetheless, when (nonlinear) function approximation is used, the convergence guarantees of all these algorithms remain rather elusive.   

}

 \vspace{-2mm}
\subsection{Modern Optimization-based RL Algorithms}\label{subsec:optimization-based-RL} \vspace{-2mm}
Leveraging the optimization perspectives of RLs, 
recent works (see, e.g., \cite{sutton2009fast,chen2016stochastic,dai2017learning,dai2018sbeed,lee2018stoch,lee2019target}) generate new principles for solving RL problems as we transition from linear towards nonlinear function approximations as well as establish theoretical guarantees based on rich theory in mathematical optimization literature.

To build up an understanding, we first recall the linear programming (LP) formulation of the \textit{planning} problem~\cite{puterman2014markov}
\begin{align}
&\min_V \quad \mu^T V\quad {\rm subject~to}\quad R_a + \gamma P_a V \le V,\quad \forall a \in {\cal A}, \label{eq:lp-planning}
\end{align}
where $\mu$ is the initial state distribution, $R_a \in {\mathbb R}^{|{\cal S}|}$ is the expected reward vector, and $P_a \in {\mathbb R}^{|{\cal S}|\times |{\cal S}|}$ is the state transition probability matrix given action $a$. The constraints in this LP naturally arise from the Bellman equations. It is known that the solution to \eqref{eq:lp-planning} is the optimal state-value function $V^*$, and that the solution to the dual of \eqref{eq:lp-planning} yields the optimal policy. By exploiting the Lagrangian duality, the optimal value function and optimal policy can be found through solving the min-max problem: 
\begin{align}
&\min_{V\in\mathcal{V}} \max_{\lambda =(\lambda_a)_{a\in {\cal A}} \in \Lambda} L(V,\lambda ):=\mu^T V + \sum_{a\in {\cal A}}{\lambda_a^T (R_a + \gamma P_a V-V)},\label{eq:saddle-point-problem}
\end{align}
where sets $\mathcal{V}$ and $\Lambda$ are properly chosen domains that restrict on the optimal value function and policy. 

{
Building on this min-max formulation, several recent works introduce efficient RL algorithms for finding the optimal policy. For instance, the stochastic primal-dual RL (SPD-RL) in~\cite{chen2016stochastic} solves the min-max problem~\eqref{eq:saddle-point-problem} with the stochastic primal-dual algorithm 
\begin{align*}
&V_{k+1}=\Pi_{\mathcal{V}}(V_k-\gamma_k\hat \nabla_V L(V_k,\lambda_k)),\quad \lambda_{k+1}=\Pi_{\Lambda}(\lambda_k+\gamma_k \hat\nabla_\lambda L(V_k,\lambda_k)),
\end{align*}
where $\hat \nabla_V L$ and $\hat \nabla_\lambda L$ are unbiased stochastic gradient estimations, which are obtained by using samples of $(s,a,r,s')$, $\Pi_{\mathcal{V}}$ and $\Pi_{\Lambda}$ stand for the projection operators onto the sets $\mathcal{V}$ and $\Lambda$. Since these gradients are obtained based on the samples, the updates can be executed without the model knowledge.  The SPD Q-learning in~\cite{lee2018stoch} extends it to the $Q$-learning framework with off-policy learning, where the sample observations are collected from some time-varying behavior policies. The dual actor-critic in~\cite{dai2018boosting} generalizes the setup to continuous state-action MDP and exploits nonlinear function approximations for both value function and the dual policy. These primal-dual type algorithms resemble the classical actor-critic methods by simultaneously updating the value function and policy, yet in a more efficient and principled manner. 
}

Apart from the LP formulation, alternative nonlinear optimization frameworks based on the fixed point interpretation of Bellman equations have also been explored, both for policy evaluation and policy optimization. { To name a few, Baird's  residual gradient algorithm~\cite{baird1995residual}, designed for policy evaluation, aims for minimizing the mean-squared Bellman error, i.e., 
\begin{align}\min_\theta\; \text{MSBE}(\theta):={\mathbb E}_{s}[({\mathbb E}_{s'}[r(s,\pi (s))+\gamma \phi^T (s')\theta]-\phi^T(s)\theta)^2]=\min_\theta\|R_\pi + \gamma P_\pi\Phi\theta-\Phi\theta\|_D^2, \label{eq:MSBE}
\end{align}
where $R_\pi$ and $P_\pi$ are the expected reward vector and state transition probability matrix under policy $\pi$, respectively, $\Phi$ is the feature matrix, $D$ is a diagonal matrix with diagonal entries being the stationary state distributions, and $\|x\|_D := \sqrt{x^T D x}$. The gradient TD (GTD)~\cite{sutton2009fast} solves the projected Bellman equation, $\Phi\theta = \Pi(R_\pi +\alpha P_\pi\Phi\theta)$, by minimizing the mean-square projected Bellman error, 
\begin{align}
&\min_\theta\;\text{MSPBE}(\theta):=  \left\| \Pi (R_\pi +\gamma P_\pi\Phi\theta)-\Phi\theta \right\|_D^2,\label{eq:MSBPE}
\end{align}
where $\Pi$ is the projection onto the range of the feature matrix $\Phi$. This is largely driven by the fact that most temporal-difference learning algorithms converge to the minimum of MSPBE. However, directly minimizing these optimization objectives (\ref{eq:MSBE}) and (\ref{eq:MSBPE}) can be challenging due to the double sampling issue and computational burden for the projections. Here, the double sampling issue means the requirement of double samples of the next stats from the current state to obtain an unbiased stochastic estimate of gradients of the objective mainly due to its quadratic nonlinearity. Alternatively,~\cite{mahadevan2014proximal,dai2017learning} get around this difficulty by resorting to min-max reformulations of the MSBE and MSBPE and introduce primal-dual type methods for policy evaluation with finite sample analysis. Similar ideas have also been employed for policy optimization based on the (softmax) Bellman optimality equation; see, e.g.,~\cite{dai2018sbeed} (called Smoothed Bellman Error Embedding (SBEED) algorithm).  


Compared to the classical RL approaches, the optimization-based RLs exhibit several key advantages. First, in many applications such as robot control, the agents' behaviors are required to mediate among multiple different objectives. Sometimes, those objectives can be formulated as constraints, e.g., safety constraints. 
In this respect, optimization-based approaches are more extensible than the traditional dynamic programming-based approaches when dealing with policy constraints. Second, existing optimization theory provides ample opportunities in developing convergence analysis for RLs with and without function approximations; see, e.g.,~\cite{chen2016stochastic,dai2018sbeed}. More importantly, these methods are highly generalizable to the multi-agent RL setup with decentralized rewards, when integrated with recent fruitful advances made in distributed optimization. This last aspect is our main focus in this survey.
}

\section{From single-agent to multi-agent RLs}\label{sec:multi-agent-RL-centralized-reward}

Cooperative MARL extends the single-agent RL to $N$ agents,  ${\cal V}=\{1,2,\ldots,N \}$, where the system's behavior is influenced by the whole team of simultaneously and independently acting agents in a common environment. This can be further classified into MARLs with centralized rewards and decentralized rewards.
\subsection{MARL with Centralized Rewards}
We start with MARLs with centralized rewards, where all agents have access to a central reward. In this setting, a multi-agent MDP can be characterized by the tuple, $({\cal S},\{{\cal A}^i \}_{i=1}^N,P,r,\gamma)$.
Each agent $i$ observes the common state $s$ and executes action $a^i \in {\cal A}^i$ inside its own action set ${\cal A}^i$ according to its local policy $\pi^i:{\cal S} \to {\cal A}^i$. The joint action $a:= (a^1,a^2,\ldots,a^N) \in {\cal A}:={\cal A}^1\times \cdots \times {\cal A}^N$ causes the state $s\in {\cal S}$ to transit to $s' \in {\cal S}$ with probability $P(s'|s,a)$, and the agent receives the common reward $r(s,a)$. The goal for each agent is to learn a local policy $\pi_*^i:{\cal S} \to {\cal A}^i,i\in {\cal V}$ such that $(\pi_*^1,\pi_*^2,\ldots,\pi_*^N)=:\pi^*$ is an optimal central policy.

Suppose each agent $i\in {\cal V}$ receives the central reward $r$ and knows the joint state and action pair $(s,a) \in {\cal S} \times {\cal A}$ (i.e., agents are JALs). Cooperative MARL, in this case, is straightforward because all agents have full information to find an optimal solution. As an example, a naive application of the Q-learning~\cite{claus1998dynamics} to multi-agent settings is
\begin{align*}
&Q_{k+1}^i(s_k,a_k)=Q_k^i(s_k,a_k) + \alpha_k \left\{ r(s_k,a_k)+\gamma \max_{a\in {\cal A}} Q_k^i(s_{k+1},a) - Q_k^i(s_k,a_k) \right\},
\end{align*}
where each agent keeps its local Q-function $Q^i: {\cal S} \times {\cal A} \to {\mathbb R}$. In particular, it is equivalent to the single-agent Q-learning executed by each agent in parallel, and $Q^i_{k} \to Q^*$ as $k \to \infty$ almost surely for all $i \in {\cal V}$; thereby $\pi_k^i(\cdot) = \argmax_a Q_k^i(\cdot,a) \to \pi_*^i (\cdot)$. 
Similarly, the policy search methods and actor-critic methods can be easily generalized to MARL with JALs~\cite{peshkin2000learning}. In such a case, coordination among agents is unnecessary to learn the optimal policy.  However, in practice, each agent may not have access to the global rewards due to limitations of communication or privacy issues; as a result, coordination protocols are essential for achieving the optimal policy corresponding to the global reward.  

 {
\subsection{Networked MARL with Decentralized Reward}
The main focus of this survey is on MARLs with decentralized rewards, where each agent only receives a local reward, and the central reward function is characterized as the average of all local rewards. The goal of each agent is to cooperatively find an optimal policy corresponding to the central reward by sharing local learning parameters over a communication network.

More formally, a coordinated multi-agent MDP with a communication network (i.e., networked MA-MDP) is given as the tuple, $({\cal S},\{ {\cal A}^i \}_{i=1}^N,P,\{ r^i \}_{i=1}^N ,\gamma,{\cal G})$, where $r^i(s,a)$ is the random reward of agent $i$ given action $a$ and the current state $s$, and $\cal G = ({\cal V},{\cal E})$ is an undirected graph (possibly time-varying or stochastic) characterizing the communication network. Each agent $i$ observes the common state $s$, executes action $a^i \in {\cal A}^i$ according to its local policy $\pi^i:{\cal S} \to {\cal A}^i$, receives the local reward $r^i(s,a)$, and the joint action $a:= (a^1,a^2,\ldots,a^N)$ causes the state $s\in {\cal S}$ to transit to $s' \in {\cal S}$ with probability $P(s'|s,a)$. The central reward is defined as  $r=\frac{1}{N}\sum_{i=1}^N {r^i}$.  In the course of learning, each agent receives learning parameters $\{\theta^j\}_{j\in {\cal N}_i}$ from its neighbors of the communication network. The overall model is illustrated as in~\cref{fig:diagram}.

\begin{figure}[H]
\begin{center}
\includegraphics[scale=0.4]{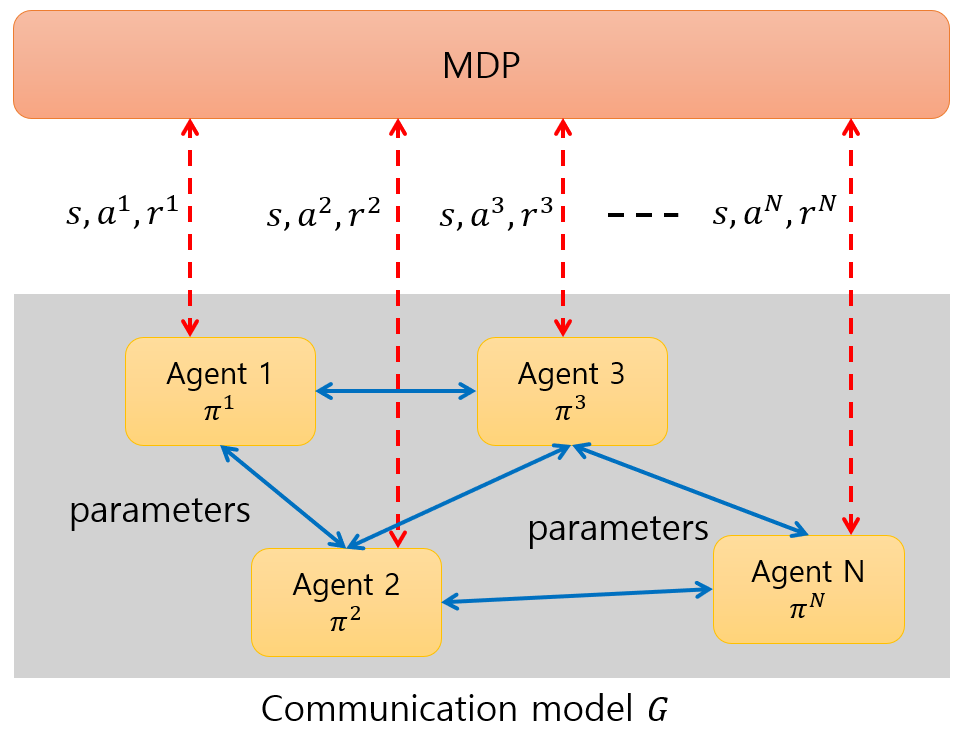}
\caption{Coordinated multi-agent MDP with communication network}\label{fig:diagram}
\end{center}
\end{figure}



For an illustrative example, we consider a wireless sensor network (WSN)~\cite{liang2008multi}, where data packets are routed to the destination node through multi-hop communications. The WSN is represented by a graph with $N$ nodes (routers), and edges connecting nodes whenever two nodes are within the communication range of each other. The route's QoS performance (quality of service) depends on the decisions of all nodes. Below we formulate the WSN as a networked MA-MDP.

\begin{example}[WSN as a networked MA-MDP]\label{ex:multi-robot-warehouse}
 
The WSN is a multi-agent system, where sensor nodes are agents. Each agent takes action $a^i\in\mathcal{A}$, which consists of forwarding a packet to one of its neighboring node $j\in {\cal N}_i$, sending an acknowledgment message (ACK) to the predecessor, dropping the data packet, where ${\cal N}_i$ is the set of neighbors of the node $i$. The global state $s=(s^{1},s^{2},\ldots,s^{N})$ is a tuple of local states $s^{i}$, which consists of the set of $i$’s neighboring nodes, and the set of packets encapsulated with QoS requirement. A simple example of the reward is $r(s,a):=\sum_{i=1}^N {r^i(s^i,a^i)}$, where
\begin{align*}
&r^i(s^i,a^i):= \begin{cases}
   1\quad {\rm if\,\,ACK\,\,received}  \\
   0\quad {\rm otherwise}  \\
\end{cases}
\end{align*}
The reward measures the quality of local routing decisions in terms of meeting with QoS requirements. 
\begin{figure}[!t]
\begin{center}
\includegraphics[scale=0.6]{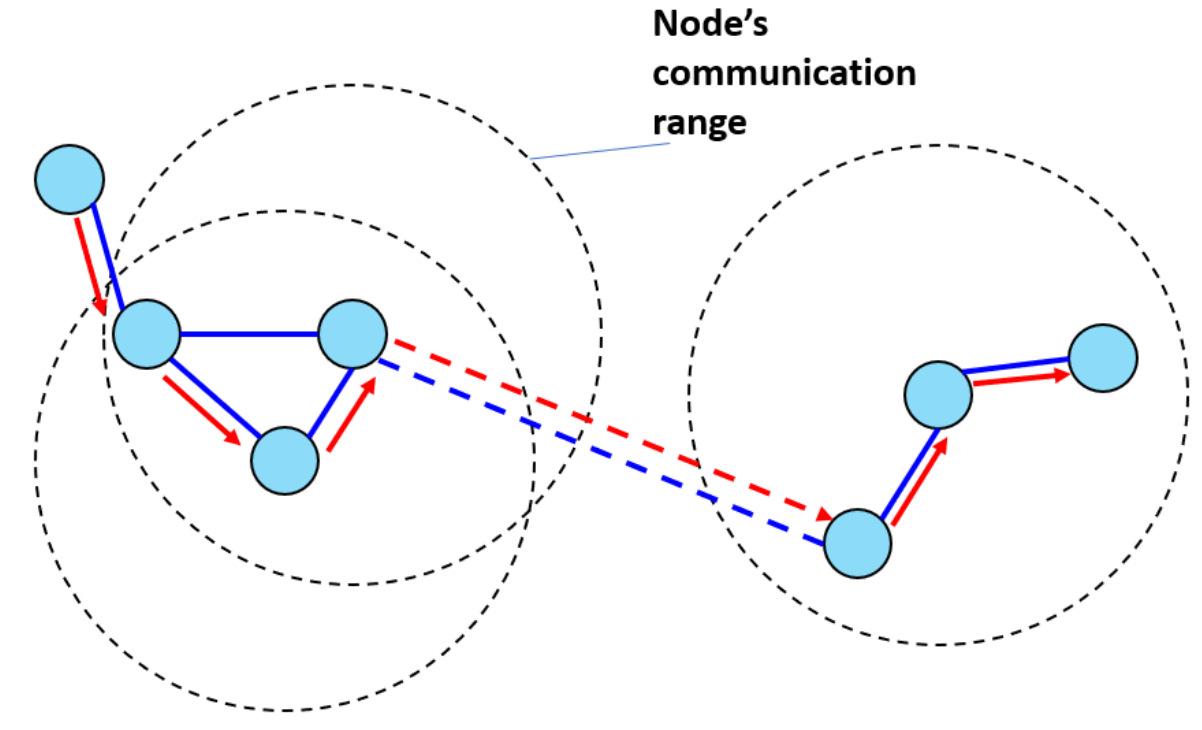}
\caption{Routing protocol for wireless sensor networks}\label{fig:network-example}
\end{center}
\end{figure}
Each agent only has access to its own reward, which measures the quality of its own routing decisions based on the QoS requirements, while the efficiency of overall tasks depends on a sum of local rewards. If each node knows the global state and action $(s,a)$, then the overall system is a networked MA-MDP.

\end{example}}

Finding the optimal policy for networked MA-MDPs naturally relates to one of the most fundamental problems in decentralized coordination and control, called the consensus problem. In the sequel, we first review the recent advances in distributed optimization and consensus algorithms, and then march forward to the discussions of recent  developments for cooperative MARL based on consensus algorithms. 


\section{Distributed optimization and consensus algorithms}\label{sec:basics-graph-consensus}


In this section, we briefly introduce several fundamental concepts in distributed optimization, which are the backbone of distributed MARL algorithms to be discussed.  


\subsection{Consensus}

Consider a set of agents, ${\cal V}=\{1,2,\ldots,N \}$, each with some initial values, $x^i(0) \in {\mathbb R}^n$. The agents are interconnected over an underlying communication network characterized by a graph ${\cal G}=({\cal V},{\cal E})$, where ${\cal E} \subset {\cal V} \times {\cal V}$ is a set of undirected edges, and each agent has a local view of the network, i.e., each agent $i\in {\cal V}$ is aware of its immediate neighbors, ${\cal N}_i$, in the network, and communicates with them only.

The goal of the consensus problem is to design a distributed algorithm that the agents can execute locally to agree on a common value as they refine their estimates. The algorithm must be local in the sense that each agent performs its own computations and communicates with its immediate neighbors only. Formally speaking, the agents are said to {\em reach a consensus} if
\begin{align}
&\lim_{k\to\infty } x^i(k)=c,\quad \forall i \in {\cal V},\label{eq:consensus-definition}
\end{align}
for some $c \in {\mathbb R}^n$ and for every set of initial values $x^i(0) \in {\mathbb R}^n$.  For ease of notation, we consider the scalar case, $n=1$, from now on.

A popular approach to the consensus problem is the \emph{distributed averaging consensus algorithm}~\cite{jadbabaie2003coordination}
\begin{align}
&x^i(k+1)=\frac{1}{|{\cal N}_i|+1}\sum_{j\in {\cal N}_i \cup \{ i\}}{x^j(k)},\quad \forall k \ge 0. \label{eq:averaging-consensus-update}
\end{align}
The averaging update is executed by local agent $i$, as it only receives values of its neighbors, $x^j(k),j\in {\cal N}_i$, and is known to ensure consensus provided that the graph is connected. Note that an undirected graph ${\cal G}$ is connected if there is a path connecting every pair of two distinct nodes. Using matrix notations, we can compactly represent \eqref{eq:averaging-consensus-update}  as follows
\begin{align}
&x(k+1)=W x(k),\quad \forall k \ge 0,\label{eq:averaging-consensus-update-compact}
\end{align}
where $x(k)$ is a column vector with entries, $x^i(k),i = 1,2,\ldots,N$, and $W$ is the weight matrix associated with~\eqref{eq:averaging-consensus-update} such that $[W]_{ij}:= \frac{1}{|{\cal N}_i|+1}$ if $j \in {\cal N}_i \cup \{ i \}$ and zero otherwise. Here, $[W]_{ij}$ means the element in the $i$-th row and $j$-th column of the matrix $W$.

The matrix $W$ is a stochastic matrix, i.e., it is nonnegative, and its row sums are one. Hence, $W^k$ converges to a rank one stochastic matrix, i.e., $\lim_{k\to \infty} W^k = {\bf 1}_n v^T$, where $v$ is the unique (normalized) left-eigenvector of $W$ for eigenvalue $1$ with $\|v\|_1 = 1$ and ${\bf 1}_n$ is an $n$-dimensional vector with all entries equal to one. Since $x(k) = W^k x(0),\forall k \ge 0$, we have $\lim_{k\to\infty} x(k) = (v^T x(0)){\bf 1}_n$, implying the consensus.


\subsection{Distributed optimization with averaging consensus}\label{sec:distributed-optimization}

Consider a multi-agent system connected over a network, where each agent $i$ has its own (convex) cost function, $f_i:{\mathbb R}^n \to {\mathbb R}$. Let $F(x):=\sum_{i\in {\cal V}} {f_i(x)}$ be the system objective that the agents want to minimize collectively. The distributed optimization problem is to solve the following optimization problem:
\begin{align}
&\min_{x\in {\mathbb R}^n} F(x):=\sum_{i=1}^{N} {f_i(x)}\quad {\rm subject}\,\,{\rm to}\quad x \in {\cal X},
\end{align}
where ${\cal X} \subseteq{{\mathbb R}^n}$ represents additional constraints on the variable $x$. By introducing local copies $x^1,x^2,\ldots,x^N$, it is equivalently expressed as
\begin{align}
&\min_{x^1\in {\cal X},\cdots,x^N\in {\cal X} } F(x):=\sum_{i=1}^{N} {f_i(x^i)}\quad {\rm subject}\,\,{\rm to}\quad  x^1 = x^2 = \cdots = x^N.
\label{eq:distributed-optimization}
\end{align}

The distributed averaging consensus algorithm can be generalized to solve the distributed optimization. An example is the \emph{consensus-based distributed subgradient method}~\cite{nedic2010constrained}, where each agent $i$ updates its local variable $x^i(k)$ according to
\begin{align*}
{\bf Consensus\,\,\, step:}\quad & w^i_{k+1}= \frac{1}{|{\cal N}_i|+1}\sum_{j\in {\cal N}_i \cup \{ i\}}{x^j_k},\\
{\bf Subgradient\,\,\, descent\,\,\, step:}\quad & x^i_{k+1}=\Pi_{{\cal X}} [w^i_{k+1}-\alpha_k \partial f_i (w^i_{k+1})],
\end{align*}
where $\partial f_i$ is any subgradient of $f_i$ and $\Pi_{{\cal X}}$ is the Euclidean projection onto the constraint set ${{\cal X}}$.

The algorithm is a simple combination of the averaging consensus and the classical subgradient method. As in the averaging consensus, the update is executed by local agent $i$, and it only receives the values of its neighbors, $x^j_k,j\in {\cal N}_i$. When all cost functions are convex, it is known that local variables, $x^i_k$, reach a consensus and converge to a solution to~\eqref{eq:distributed-optimization}, $x^*\in {\cal X}$, under properly chosen step-sizes.

Other distributed optimization algorithms include the EXTRA~\cite{shi2015extra} {(exact first-order algorithm for decentralized consensus optimization)}, push-sum algorithm~\cite{nedic2014distributed} for directed graph models, gossip-based algorithm~\cite{nedic2010asynchronous}, and etc. A comprehensive and detailed summary of the distributed optimization can be found in the monograph~\cite{nedich2015convergence}.


{
\subsection{Distributed min-max optimization with averaging consensus}\label{sec:distributed-primal-dual}

To put it one step further, distributed averaging consensus algorithm can also be generalized to solve the min-max problem in a distributed fashion. The distributed min-max optimization problem deals with the zero-sum game: 
\begin{align}
&\min_{x \in {\cal X}}\max_{\lambda\in \Lambda}L(x,\lambda):=\sum_{i = 1}^N {L^i(x,\lambda)}, 
\end{align}
where $L:{\mathbb R}^n\times {\mathbb R}^m \to {\mathbb R}$ is a convex-concave function and $L$ is separable. By introducing local copies $x^1,x^2,\ldots,x^N$, $\lambda^1,\lambda^2,\cdots,\lambda^N$, the min-max problem is equivalently expressed as 
\begin{align}
&\min_{x^1,\ldots,x^N \in {\cal X}}\max_{\lambda^1,\ldots,\lambda^N\in \Lambda}\sum_{i = 1}^N {L^i(x^i,\lambda^i)}\quad {\rm s.t.}\quad x^1 =x^2 =\cdots=x^N ,\quad \lambda^1 =\lambda^2 =\cdots=\lambda^N.\label{eq:distributed-saddle-point-problem}
\end{align}
Similar to the distributed subgradient method, the distributed primal-dual algorithm works by performing averaging consensus and sugradient descent for the local variable $x^i(k)$ and $\lambda^i(k)$ of each agent:
\begin{align*}
{\bf Consensus\,\,\, step:}
& \quad x^i_{k+1/2}= \frac{1}{|{\cal N}_i|+1}\sum_{j\in {\cal N}_i \cup \{ i\}}{x^j_k},\quad \lambda^i_{k+1/2}= \frac{1}{|{\cal N}_i|+1}\sum_{j\in {\cal N}_i \cup \{ i\}}{\lambda^j_k},\\
{\bf Primal\text{-}dual\,\,\, step:}
&\quad x^i_{k+1}=\Pi_{{\cal X}} [x^i_{k+1/2}-\alpha_k \partial_x L_i (x^i_{k+1/2},\lambda^i_{k+1/2})],\\
&\quad \lambda^i_{k+1}=\Pi_{\Lambda} [\lambda^i_{k+1/2}-\beta_k \partial_{\lambda} L_i (x^i_{k+1/2},\lambda^i_{k+1/2})]
\end{align*}
where $\alpha_k$ and $\beta_k$ are step-sizes, $\partial_x L_i$ and $\partial_\lambda L_i$ are any subgradients of $L_i(x,\lambda)$ with respect to $x$ and $\lambda$, respectively, and $\Pi_{{\cal X}}$ and $\Pi_{\Lambda}$ are the Euclidean projection onto the constraint sets ${\cal X}$ and $\Lambda$, respectively. The distributed primal-dual algorithm and other variants have been well studied in~\cite{zhu2011distributed,hong2017prox,yuan2011distributed}. 

}

\section{Networked MARL with decentralized rewards}\label{sec:multi-agent-RL-decentralized-rewards}

In this section, we focus on networked MARL with decentralized rewards, where the corresponding networked MA-MDP is described by the tuple, $({\cal S},\{ {\cal A}^i \}_{i=1}^N,P,\{ r^i \}_{i=1}^N ,\gamma,{\cal G})$.  The goal of each agent is to cooperatively find an optimal policy corresponding to the central reward, $r=(r^1+r^2+\cdots +r^N)/N$, by sharing local learning parameters over a communication network characterized by graph ${\cal G}=({\cal V},{\cal E})$.

Decentralized rewards are common in practice when multiple agents cooperate to learn under sensing and physical limitations.  Consider multiple robots navigating and executing multiple tasks in geometrically separated regions. The robots receive different rewards based on the space they reside in.   Decentralized rewards are also particularly useful when MARL agents cooperate to learn an optimal policy securely due to privacy considerations. For instance, if we do not want to reveal full information about the policy design criterion to an RL agent to protect privacy, a plausible approach is to operate multiple RL agents, and provide each agent with only partial information about the reward function. In this case, no single agent alone can learn the optimal policy corresponding to the whole environment, without information exchange among other agents. 
Most recent algorithms to be discussed in this section, including~\cite{doan2019convergence,wai2018multi,lee2018stochastic,macua2015distributed,stankovic2016multi,kar2013cal,zhang2018fully,zhang2018finite,qu2019value}, apply  the distributed averaging consensus algorithm introduced in~\cref{sec:basics-graph-consensus} in one way or another. We now discuss these algorithms in details below, with a brief summary provided in Table~\ref{table:Coordinated}.

\begin{table}[t]
\caption{Cooperative MARL with decentralized rewards and communication networks (LFA: linear function approximation; NFA: nonlinear function approximation; N/A: Not Applicable}\label{table:Coordinated}
\begin{center}
\begin{tabular}{p{2.4cm}|p{2.6cm}|p{1.5cm}|p{1.7cm}|p{1.4cm}|p{1.8cm}}
\hline
& Papers & Availability of actions & Reward & Function Approx. &Convergence  \\
\hline
\multirow{5}*[0ex]{Policy Evaluation} & Doan et al.\cite{doan2019convergence} & \multirow{3}*[0ex]{N/A} & \multirow{3}*[0ex]{Decentralized} & LFA  & Yes\\
& Wai et al.\cite{wai2018multi} & & & LFA   & Yes\\
& Lee \cite{lee2018stochastic} & &  & LFA   & Yes\\
\cline{2-6}
& Macua et al.\cite{macua2015distributed} & \multirow{2}*[0ex]{N/A} & \multirow{2}*[0ex]{Centralized} & LFA  & Yes \\
& Stankovi{\'c} et al.\cite{stankovic2016multi} &  & & LFA & Yes \\

\hline\hline

\multirow{4}*[0ex]{Policy Optimization} & Kar et al.\cite{kar2013cal} & JAL & \multirow{4}*[0ex]{Decentralized} & Tabular  & Yes\\
& Zhang et al.\cite{zhang2018fully} & JAL & & LFA, NFA & Yes \\
& Zhang et al.\cite{zhang2018finite} & JAL & & LFA, NFA & Local \\
& Qu et al.\cite{qu2019value} & JAL & & NFA & Local  \\

\hline
\end{tabular}
\end{center}
\end{table}

\subsection{Distributed Policy Evaluation}\label{sec:multi-agent-RL-decentralized-rewards-eval}

The goal of distributed policy evaluation is to evaluate the central value function
\begin{align*}
&V^\pi(s) = {\mathbb E}\left[ \left. \sum_{k = 0}^\infty  {\gamma ^k \frac{1}{N}\sum_{i=1}^N {r_\pi^i(s_k)} } \right|s_0=s\right]
\end{align*}
in a distributed manner. The information available to each agent is $(s,r^i,\{\theta^j\}_{j\in {\cal N}_i})$, where $\{\theta^j\}_{j\in {\cal N}_i}$ represents the set of learning parameters agent $i$ receives from its neighbors over the communication network, and ${\cal N}_i$ is the set of all neighbors of node $i$ over the graph ${\cal G}$. Note that for policy evaluation with state value function $V$, the information $a$ or $a^i$ is not necessary, thereby it is not indicated in the information set $(s,r^i,\{\theta^j\}_{j\in {\cal N}_i})$.

The distributed TD-learning~\cite{doan2019convergence} executes the following local updates of agent $i$:
\begin{align*}
&\theta^i \leftarrow \underbrace{\frac{1}{|{\cal N}_i|+1} \sum_{j \in {\cal N}_i \cup \{ i\} } {\theta^j}}_{\rm Mixing\,\,term} +\underbrace{\gamma (r^i(s,\pi(s))+\gamma\phi(s')^T \theta^i  -\phi(s)^T \theta^i)\phi(s)}_{\rm TD\,\,update},
\end{align*}
where each agent $i$ keeps its local parameter $\theta^i$. The algorithm resembles the consensus-based distributed subgradient method in~\cref{sec:distributed-optimization}. The first term, dubbed as the mixing term, is an average of local copies of the learning parameter of neighbors, ${\cal N}_i$, received from communication over networks, and controls local parameters to reach a consensus. The second term, referred to as the TD update, follows the standard TD updates. Under suitable conditions such as the graph connectivity, each local copy, $\theta^i$, converges to $\theta^*$ in expectation and almost surely~\cite{doan2019convergence}, where $\theta^*$ is the optimal solution found by the single-agent TD learning acting on the central reward. 

\subsection{Distributed Policy Optimization}\label{sec:multi-agent-RL-decentralized-rewards-impro}

The goal of distributed policy optimization is to cooperatively find an optimal central policy corresponding to the central reward, $r$. Note that the distributed TD-learning in the previous section only finds the state value function under a given policy. The averaging consensus idea can also be extended to Q-learning and actor-critic algorithms for finding the optimal policy for networked MARL. 

\begin{figure}[H]
\begin{center}
\includegraphics[scale=0.45]{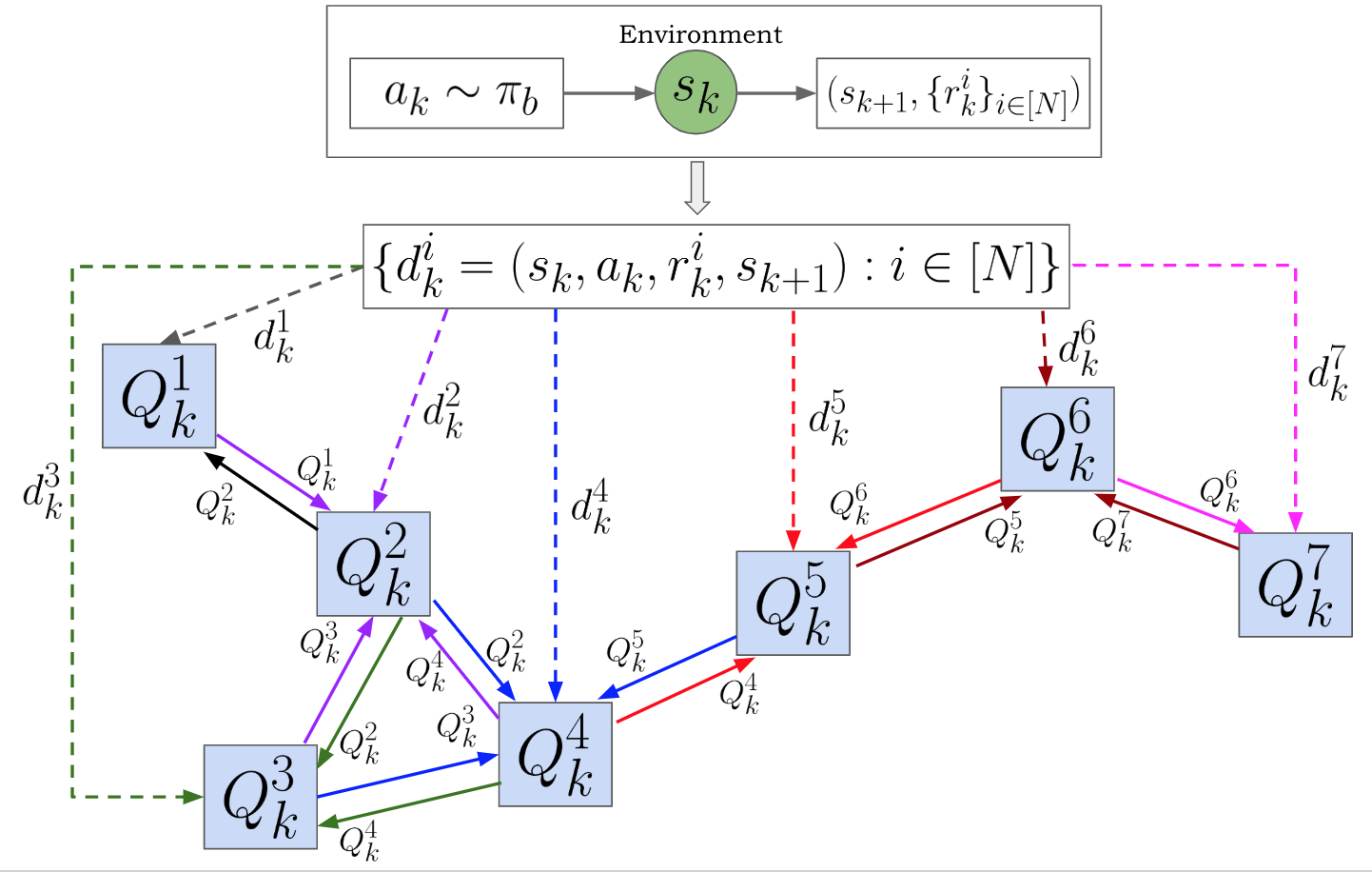}
\caption{Diagram of distributed Q-learning algorithm in~\cite{kar2013cal}. Here the joint-action $a_k$ is chosen by a behavior policy $\pi_b$.}\label{fig:distributed-Q-learning}
\end{center}
\end{figure}

The distributed Q-learning in~\cite{kar2013cal} locally updates the Q-function according to
\begin{align*}
Q^i(s,a) \leftarrow& Q^i(s,a) - \eta(s,a) \underbrace{\sum_{j \in {\cal N}_i \cup \{ i\} } {(Q^i(s,a)-Q^j (s,a))}}_{\rm Mixing\,\, term}  \\
&+\alpha(s,a)\underbrace{(r^i(s,a)+\gamma \max_{a'\in {\cal A}} Q^i(s',a')- Q^i(s,a))}_{\rm Q-learning\,\, update},
\end{align*}
where $i$ is the agent index, $\eta(s,a)$ and $\alpha(s,a)$ are learning rates (or step-sizes) depending on the number of instances when $(s,a)$ is encountered. The information available to each agent is $(s,a,r^i,\{Q^j\}_{j\in {\cal N}_i\cup \{i\}})$. The overall diagram of the distributed Q-learning algorithm is given in~\cref{fig:distributed-Q-learning}. Each agent $i$ keeps the local Q-function, $Q^i$, and the mixing term consists of Q-functions of neighbors received from communication networks. It has been shown that each local $Q^i$ reaches a consensus and converges to $Q^*$ almost surely~\cite{kar2013cal} with suitable step-size rules and under assumptions such as the connectivity of the graph and an infinite number of state-action visits. 

\begin{figure}[H]
\begin{center}
\includegraphics[scale=0.6]{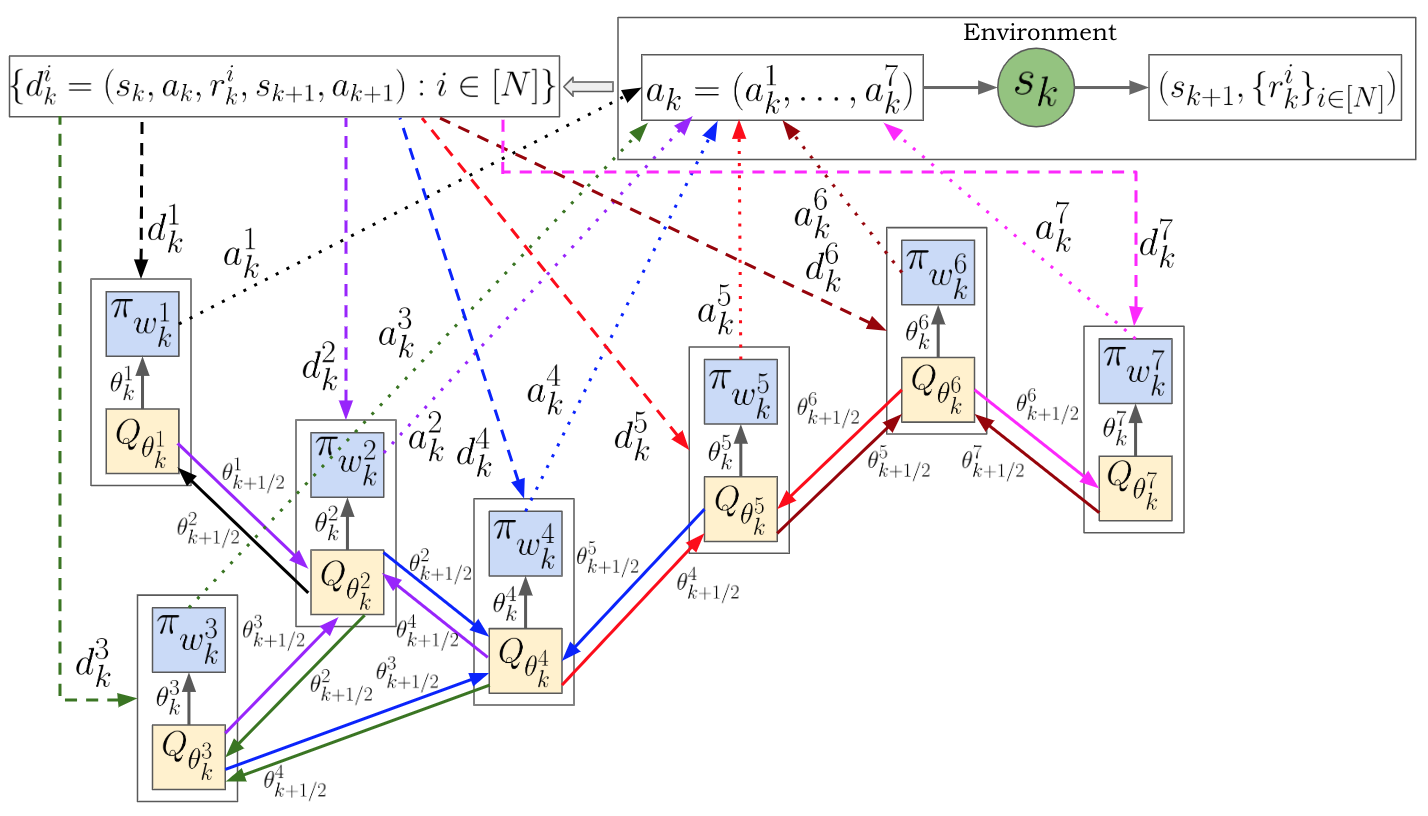}
\caption{Diagram of distributed actor-critic algorithm in~\cite{zhang2018fully}. Here the joint-action $a_k$ is taken in on-policy manner. }\label{fig:distributed-actor-critic}
\end{center}
\end{figure}

The distributed actor-critic algorithm in~\cite{zhang2018fully} generalizes the single-agent actor-critic to networked MA-MDP settings where the averaging consensus steps are taken for the value function parameters 
{\begin{align*}
&{\rm Critic\,\, update}:\theta_{k+1/2}^i=\theta_k^i + \alpha_k (r^i(s_k,a_k)+\gamma Q(s_{k+1},a_{k+1};\theta_k^i)-Q(s_k,a_k;\theta_k^i))\nabla_\theta Q(s_k,a_k;\theta_k^i)\\
&{\rm Actor\,\, update}:
w_{k+1}^i=w_k^i + \beta_k A(s_k,a_k;\theta_k^i)\nabla_{w^i} \log\pi_{w_k^i}^i(s_k,a_k^i)\\
&{\rm Mixing\,\, step}:\theta_{k+1}^i=\frac{1}{|{\cal N}_i|+1}\sum_{j \in {\cal N}_i \cup \{i\}} {\theta_{k+1/2}^j }
\end{align*}
where $w^i$ and $\theta^i$ are parameters of nonlinear function approximations for the local actor and local critic, respectively. Here $A(s_k,a_k;\theta_k^i):=Q(s_k,a_k;\theta_k^i)-\sum_{a^i \in {\cal A}^i} {\pi_{w_k^i}^i (s_k,a^i)Q(s_k,(a_k^1,\ldots,a^i,\ldots,a_k^N);\theta_k^i)}$ is the advantage function evaluated at $(s_k,a_k)$.}
The overall diagram of the distributed actor-critic is given in~\cref{fig:distributed-actor-critic}. Each agent $i$ keeps its local parameters $\{\theta^i,w^i\}$, and in the mixing step, it only receives local parameters of the critic from neighbors. The actor and critic updates are similar to those of typical actor-critic algorithms with local parameters. The information available to each agent is $(s,a,r^i,w^i,\{\theta^j\}_{j\in {\cal N}_i \cup \{i\}})$. 
The results in~\cite{zhang2018finite} study a MARL generalization of the fitted Q-learning with the information structure $(s,a,r^i,\{\theta^j\}_{j\in {\cal N}_i\cup \{i\} })$. { Compared to the tabular distributed Q-learning in~\cite{kar2013cal}, the distributed actor-critic and fitted Q-learning may not converge to an exact optimal solution mainly due to the use of function approximations. }

\subsection{Optimization Frameworks for Networked MA-MDP}\label{subsec:optimization-MA-MDP}
Recall that in \cref{subsec:optimization-based-RL}, we discussed optimization frameworks of single-agent RL problem. By integrating them with consensus-based distributed optimization, they can be naturally  adapted to solve networked MA-MDPs. In this subsection, we introduce some recent work in this direction, such as the value propagation~\cite{qu2019value}, primal-dual distributed incremental aggregated gradient~\cite{wai2018multi}, distributed GTD~\cite{lee2018stochastic}. The main idea of these algorithms is essentially rooted in formulating the overall MDP into a min-max optimization problem, $\min_{x \in {\cal X}}\max_{\lambda\in \Lambda}L(x,\lambda)$, with separable function $L(x,\lambda) = \sum_{i = 1}^N {L^i(x,\lambda)}$, and solving the distributed min-max optimization problem~\eqref{eq:distributed-saddle-point-problem}. For MARL tasks, the distributed min-max problem can be solved using stochastic variants of the distributed saddle-point algorithms in~\cref{sec:distributed-primal-dual}.  

{
The multi-agent policy evaluation algorithms in~\cite{wai2018multi} and~\cite{lee2018stochastic}  are multi-agent variants of the GTD~\cite{sutton2009fast} based on the consensus-based distributed saddle-point framework for solving the mean-squared projected Bellman error in~\eqref{eq:MSBPE},
which can be equivalently converted into an optimization problem with separable objectives:
\begin{align}
&\min_{\theta} \frac{1}{2}\sum_{i=1}^N{\|\Pi(R_\pi^i+\alpha P^\pi\Phi\theta)-\Phi\theta\|_D^2}.\label{eq:separable-MSPBE}
\end{align}
To alleviate the double sampling issues in GTD, the approach in~\cite{wai2018multi} applies the Fenchel duality with an additional proximal term to each objective, arriving at the reformulation:
\begin{align*}
&\min_{\{\theta^i\}_{i=1}^N}\sum_{i=1}^N {d^i(\theta^i)}\quad {\rm s.t.}\quad \theta^1=\theta^2=\cdots=\theta^N,
\end{align*}
where the local objectives are expressed as max-forms
\begin{align*}
&d^i(\theta):=\max_{w^i}\{J_i(\theta,w_i):=w_i^T (\Phi^T D((1/N)R_\pi^i+\alpha P^\pi\Phi\theta)-\Phi\theta)-(1/2)w_i^T \Phi^T D\Phi w_i +(\rho/2)\|\theta^i \|_2^2\} .
\end{align*}
The resulting problem can be solved by using stochastic variants of the consensus-based distributed subgradient method akin to~\cite{qu2017harnessing}. In particular, the algorithm introduces gradient surrogates of the objective function with respect to the local primal and dual variables, and the mixing steps for consensus are applied to both the local parameters and local gradient surrogates. The main idea of the primal-dual algorithm used in~\cite{qu2017harnessing} is briefly (with some simplifications) written by 
\begin{align*}
&{\rm Primal\,\,update}:\theta_{k+1}^i=\underbrace {\frac{1}{|{\cal N}_i|+1}\sum_{j\in {\cal N}_i\cup \{ i\} } {\theta_k^j}}_{\rm mixing\,\,term}-\alpha \hat g_k^i\\
&{\rm Dual\,\,update}:w_{k+1}^i=w_k^i+ \beta \hat h_k^i 
\end{align*}
where $\alpha$ and $\beta$ are step-sizes, $\hat g^i_k$ and $\hat h^i_k$ are surrogates of the gradients, $\nabla_{\theta^i} J_i(\theta_k^i,w_k^i)$ and $\nabla_{w^i} J_i(\theta_k^i,w_k^i)$, respectively, from through some basic gradient tracking steps.

The multi-agent policy evaluation in~\cite{lee2018stochastic} approaches in a different way to solve~\eqref{eq:separable-MSPBE}. Assuming each parameter $\theta^i$ is scalar for simplicity, the distributed optimization~\eqref{eq:separable-MSPBE} can be converted into
\begin{align*}
&\min_{\{\theta^i\}_{i=1}^N} \frac{1}{2}\sum_{i=1}^N {\|\Pi(R_\pi^i+\alpha P^\pi\Phi\theta^i)-\Phi\theta^i\|_D^2}+\bar\theta^T L^T L\bar\theta \quad {\rm s.t.}\quad L\bar\theta= 0,
\end{align*}
where $\bar\theta$ is the vector enumerating the local parameters, $\{\theta^i\}_{i=1}^N$, and $L=L^T \in {\mathbb R}^N$ is the graph Laplacian matrix. Note that if the underlying graph is connected, then $L\bar\theta= 0$ if and only if $\theta^1=\theta^2=\cdots=\theta^N$. By constructing the Lagrangian dual of the above constrained optimization, we obtain the corresponding single min-max problem. Thanks to the Laplacian matrix, the corresponding stochastic primal-dual algorithm is automatically decentralized. Compared to~\cite{qu2017harnessing}, it only needs to share local parameters with neighbors rather than the gradient surrogates.


The MARL in~\cite{qu2019value} combines the averaging consensus and SBEED~\cite{dai2018sbeed} (Smoothed
Bellman Error Embedding), which is called distributed SBEED here. In particular, the distributed SBEED aims to solve the so-called smoothed Bellman equation
 \begin{align*}
&V_{\theta}(s) = \frac{1}{N}\sum_{i=1}^N {R_a^i(s)} +\gamma {\mathbb E}_{s'\sim P(\cdot|s,a)} [V_{\theta}(s')]- \lambda\sum_{i=1}^N{\ln(\pi_{w^i}^i(s,a^i))},
\end{align*}
by minimizing the corresponding mean squared smoothed Bellman error:
\begin{align*}
&\min_{\theta,\,\{w^i\}_{i=1}^N} {\mathbb E}_{s,a} \left[ \left( \frac{1}{N}\sum_{i=1}^N {R_a^i(s)}+\gamma {\mathbb E}_{s'\sim P(\cdot |s,a)} [V_{\theta}(s')]- \lambda\sum_{i=1}^N{\ln(\pi_{w^i}^i(s,a^i))} - V_{\theta}(s) \right)^2 \right],
\end{align*}
where $\lambda$ is a positive real number capturing the smoothness level, $\theta$ and $w$ are deep neural network parameters for the value and policy, respectively. Directly applying the stochastic gradient to the above objective using samples leads to biases due to the nonlinearity of the objective (or double sampling issue). To alleviate this difficulty, the distributed SBEED introduces the primal-dual form as in~\cite{dai2018sbeed}, which results in a distributed saddle-point problem similar to~\eqref{eq:distributed-saddle-point-problem} and is processed with a stochastic variants of the distributed proximal primal-dual algorithm in~\cite{hong2017prox}.}

\subsection{Special Case: Networked MARL with Centralized Rewards}
Lastly, we remark that the algorithms in this section can be directly applied to MA-MDPs with central rewards. As in~\cref{sec:multi-agent-RL-centralized-reward}, we consider an MDP, $({\cal S},{\cal A},P,r,\gamma)$, with an additional network communication model ${\cal G}$, while each agent $i$ receives the common reward $r(s,a)$ instead of the local reward $r^i(s,a)$.  
One may imagine reinforcement learning algorithms running in $N$ identical and independent simulated environments. Under this assumption, a distributed policy evaluation was studied in~\cite{stankovic2016multi}. It combines GTD~\cite{sutton2009fast} with the distributed averaging consensus algorithm as follows:
\begin{align*}
&{\rm GTD\,\, update}: \begin{cases}
 \theta^i_{k+1/2}=\theta^i_k+\alpha_k(\phi(s)-\gamma\phi(s'))(\phi(s)^T w^i_k)\\
 w^i_{k+1/2}=w^i_k+\alpha_k(\delta^i_k-\phi(s)^T w^i_k)\phi(s)\\
 \end{cases}\\
&{\rm Mixing\,\, term}:\begin{cases}
 \theta^i_{k+1}=\frac{1}{|{\cal N}_i|+1}\sum_{j\in {\cal N}_i \cup \{i\} } {\theta^j_{k+1/2}}\\
 w^i_{k+1}=\frac{1}{|{\cal N}_i|+1}\sum_{j\in {\cal N}_i \cup \{i\}} {w^j_{k+1/2}}\\
 \end{cases}
\end{align*}
where $\delta^i_k=r(s,\pi(s))+\gamma\phi(s')^T \theta^i_k-\phi(s)^T \theta^i_k$ is the local TD-error. Each agent has access to the information $(s,a,r,\{\theta^j\}_{j\in {\cal N}_i})$, while the action $a$ is not used in the updates. The first update is equivalent to the GTD in~\cite{sutton2009fast} with a local parameter $(\theta^i,w^i)$ and the second term is equivalent to the distributed averaging consensus update in~\eqref{eq:averaging-consensus-update}. Since the GTD update rule is equivalent to a stochastic primal-dual algorithm, the above update rule is equivalent to a distributed algorithm for solving the distributed saddle-point problem in~\eqref{eq:distributed-saddle-point-problem}. Note that~\cite{stankovic2016multi} only proves the weak convergence of the algorithm. In the same vein, the multi-agent policy evaluation~\cite{macua2015distributed} generalizes the GQ learning to distributed settings, which is more general than GTD in the sense that it incorporates an importance weight of agent $i$ that measures the dissimilarity between the target and behavior policy for the off-policy learning.

{
\section{Future Directions}\label{sec:MARL-computational-cost}

 Until now, we mainly focused on networked MARL and recent advances which combine tools in consensus-based distributed optimization with MARL under decentralized rewards. There remain much more challenging agendas to be studied. By bridging two domains in a synergistic way, these research topics are expected to generate new results and enrich both fields.
 
 \paragraph{Robustness of networked MARL} Communication networks in real world, oftentimes, suffer from communication delays, noises, link failures, or packet drops. Moreover, network topologies may vary as time goes by and the information exchange over the networks may not be bidirectional in general.  Extensive results on distributed optimization algorithms over time-varying, directed graphs, w/o communication delays have been actively studied in the distributed optimization community, yet mostly in deterministic and convex settings. The study of networked MARLs under aforementioned communication limitations is an open and challenging topic. 
 

\paragraph{Resilience of networked MARL} Building resilient networked MARL under adversarial attacks is another important topic. A resilient consensus-based distributed optimization algorithm under adversarial attacks has been studied in~\cite{sundaram2018distributed}, which considers scenarios where adversarial agents exist among networked agents and send arbitrary parameters to their neighboring agents to disrupt the solution search. In such cases, analysis of fundamental limitations on distributed optimization algorithms and protocols resilient against such adversarial behaviors are available. For networked MARL, such issues remain largely unexplored.

\paragraph{Development of  deep networked MARL algorithms}  Another interesting direction is the application of consensus-based distributed optimizations to recent deep RL algorithms, such as deep Q-learning~\cite{mnih2015human}, trust region policy optimization (TRPO)~\cite{schulman2015trust}, proximal policy optimization (PPO)~\cite{schulman2017proximal}, deep deterministic policy gradient (DDPG)~\cite{lillicrap2015continuous}, twin delayed DDPG (TD3)~\cite{fujimoto2018addressing}. Most of these algorithms are variants of policy search algorithm and involve optimization procedures in certain stages.  Ideas of distributed optimizations can potentially be applied to these deep RL algorithms as well.


\paragraph{Theoretical understanding of networked MARL with deep neural nets} Fundamental analysis of  networked MARL with nonlinear function approximation is still an open question. For the optimization-based MARLs, when the value function or policy are parameterized by deep neural networks, the resulting distributed min-max problems discussed  eventually become nonconvex-nonconcave. Solving this class of distributed optimization problems in a principled manner remains an intriguing research topic.


\paragraph{MARL for parallel computing} Lastly, networked MARLs can be used to reduce memory and computational cost, and accelerate the training by exploiting parallel computation. 
Most RL algorithms require enormous experiences to find a reasonably good policy, which may not be easily collected by a single agent. Instead, a large number of cooperative RL agents over networks can more effectively collect experiences using their own sensors such as crowd sources. Moreover, these agents can learn  different parts of learning parameters and features with lower dimensions compared to the state-space, which could greatly reduce the memory and computational cost. There exist several works in this direction, such as the distributed gossiping TD-learning in~\cite{mathkar2017distributed},the distributed policy search algorithm~\cite{peshkin2000learning}, etc. In this case, the design of network topology and infrastructures becomes quite critical in improving the learning efficiency and balancing the tradeoff between communication and computation cost. 

}

\bibliographystyle{IEEEtran}
\bibliography{reference}


\end{document}